\documentclass[conference]{IEEEtran}
\IEEEoverridecommandlockouts
\usepackage{cite}
\usepackage{amsmath,amssymb,amsfonts}
\usepackage{algorithmic}
\usepackage{graphicx}
\usepackage{textcomp}
\usepackage{xcolor}
\def\BibTeX{{\rm B\kern-.05em{\sc i\kern-.025em b}\kern-.08em
    T\kern-.1667em\lower.7ex\hbox{E}\kern-.125emX}}
\usepackage{adjustbox}
\usepackage{booktabs}
\usepackage[caption=false,font=normalsize,labelfont=sf,textfont=sf]{subfig}
\usepackage{multirow}
\usepackage{hyperref}

\begin{document}

\title{Label Unbalance in High-frequency Trading}

\author{%
    \IEEEauthorblockN{Zijian Zhao \IEEEauthorrefmark{1} \IEEEauthorrefmark{2}, Xuming Zhang \IEEEauthorrefmark{1} \IEEEauthorrefmark{3}, Jiayu Wen \IEEEauthorrefmark{1} \IEEEauthorrefmark{4}, Mingwen Liu \IEEEauthorrefmark{1}, Xiaoteng Ma \IEEEauthorrefmark{1} \IEEEauthorrefmark{5} \thanks{Corresponding Email: zhaozj28@mail2.sysu.edu.cn }}%
    \IEEEauthorblockA{
    \IEEEauthorrefmark{1} Likelihood Lab 
    \IEEEauthorrefmark{2} Sun Yat-sen University   %
    \IEEEauthorrefmark{3} Peking University \\ %
    \IEEEauthorrefmark{4} The London School of Economics and Political Science %
    \IEEEauthorrefmark{5} Tsinghua University  %
    }%
}


\maketitle

\begin{abstract}
In financial trading, return prediction is one of the foundation for a successful trading system. By the fast development of the deep learning in various areas such as graphical processing, natural language, it has also demonstrate significant edge in handling with financial data. While the success of the deep learning relies on huge amount of labeled sample, labeling each time/event as profitable or unprofitable, under the transaction cost, especially in the high-frequency trading world, suffers from serious label imbalance issue. 
In this paper, we adopts rigurious end-to-end deep learning framework with comprehensive label imbalance adjustment methods and succeed in predicting in high-frequency return in the Chinese future market. 
The code for our method is publicly available at \href{https://github.com/RS2002/Label-Unbalance-in-High-Frequency-Trading}{https://github.com/RS2002/Label-Unbalance-in-High-Frequency-Trading}.
\end{abstract}


\section{Introduction}
In high-frequency trading (HFT), sophisticated algorithms make rapid trading decisions based on large volumes of financial data within milliseconds. These decisions often hinge on predictive models that identify favorable trading opportunities. 

However, a significant challenge in developing these predictive models is the issue of label imbalance, where certain outcomes or events (labels) are far less frequent than others. For instance, predicting rare events like sudden market crashes or abrupt price changes can be critical for high-frequency traders, but the scarcity of these events in historical data skews the distribution of labels. This imbalance complicates model training, as standard machine learning algorithms may become biased towards the more frequent labels, leading to poor generalization and a higher risk of substantial financial losses. Addressing label imbalance is crucial to enhance the robustness and reliability of predictive models in high-frequency trading, ultimately driving profitability and minimizing risk exposure in fast-paced market environments.

\section{Related Work}
Rare events are events that happen much less frequently compared to common events. In the field of data mining, identifying rare events is typically a classification problem. Due to their infrequency and casual nature, rare events are challenging to detect, and misclassifying them can lead to significant costs. For instance, mislabeling a return in high-frequency trading (HFT) can result in incorrect buying or selling decisions, leading to profit loss. The scarce occurrences of rare events make the detection task an imbalanced data classification problem. Imbalanced data refers to a dataset in which one or more classes have a significantly greater number of examples than the others. The most 
prevailing class is called the majority class, whereas the rarest class is called the minority class, typically representing the concept of interest. Although data mining methods are widely used to develop classification models for guiding business and managerial decision-making, the classification of imbalanced data poses significant challenges to traditional classification models. Since most standard classification algorithms, such as logistic regression, support vector machine (SVM), and decision trees, are designed for balanced training sets, they may produce suboptimal classification models, leading to good coverage of the majority examples but frequent misclassification of minority ones. As a result, algorithms that perform well in standard classification frameworks do not necessarily deliver the best performance for imbalanced datasets \cite{reviewOfImbalanced}. There are several reasons for this behaviour \cite{Lopez2013}:
\begin{enumerate}
    \item The use of global performance measures, like the standard accuracy rate, to guide the learning process may bias the results towards the majority class.
    \item Classification rules that predict the positive class are often highly specific and have very low coverage. This leads to the exclusion of the class in favour of more general rules that predict the negative class.
    \item Very small clusters of minority class examples can be identified as noise, which could be falsely discarded by the classifier. On the contrary, noise may be falsely identified as minor examples, as both are rare patterns in the sample space.
\end{enumerate}

The machine learning community has shown significant interest in addressing the imbalanced learning problem in recent years. Over the past decade, various machine learning approaches have been developed to tackle imbalanced data classification. These approaches primarily rely on preprocessing techniques, cost-sensitive learning, and ensemble methods \cite{wang2021, reviewOfImbalanced, Lopez2013}. Below, we will delve into the details of these three methods:

\subsection{Preprocessing techniques}
Preprocessing is commonly conducted before constructing a learning model to improve the quality of input data. Two classical techniques are frequently utilised as preprocessors:

\subsubsection{Resampling}
Resampling techniques are implemented to rebalance the sample space for imbalanced datasets, aiming to mitigate the impact of the skewed class distribution during the learning process. They can be categorised into three groups based on the method used to balance the class distribution \cite{reviewOfImbalanced}:

\begin{itemize}
    \item Over-sampling methods: these mitigate the adverse effects of skewed distribution by generating new minority class samples. Two widely used methods for creating synthetic minority samples are randomly duplicating the minority samples and employing the 'Synthetic Minority Oversampling Technique' (SMOTE).

    \item Under-sampling methods: these address the issues arising from skewed distribution by discarding samples from the majority class. Random Under Sampling (RUS) is the simplest yet highly effective method, which involves randomly eliminating majority class examples.

    \item Hybrid methods: these techniques combine over-sampling and under-sampling methods.
\end{itemize}

In studies comparing the effectiveness of various re-sampling methods, important insights were gained regarding the selection of re-sampling methods \cite{reviewOfImbalanced, zhou2013, loyola2016}. When dealing with datasets containing hundreds of minority observations, it was observed that an under-sampling method outperformed an over-sampling method in terms of computational time. However, in scenarios where only a few dozen minority instances were present, the over-sampling method SMOTE was found to be the preferable option. When the training sample size is excessively large, a hybrid approach involving SMOTE and under-sampling is recommended as an alternative. Finally, SMOTE exhibits a slightly higher efficacy in detecting outliers.

\subsubsection{Feature selection and extraction}

In general, the aim of feature selection is to choose a subset of k features from the entire feature space. This subset should enable a classifier to achieve optimal performance, where k is a user-specified or adaptively selected parameter. Feature selection can be categorised into filters, wrappers, and embedded methods  \cite{reviewOfImbalanced}.

Another approach to address dimensionality is feature extraction, which corresponds to dimensionality reduction and involves transforming data into a low-dimensional space. However, it's important to note that feature selection techniques are distinct from feature extraction. While feature selection returns a subset of the original features, feature extraction generates new features from the original ones using functional mapping. There are various techniques for feature extraction, such as Principal Component Analysis (PCA), Singular Value Decomposition (SVD), and Non-negative Matrix Factorization (NMF). Feature extraction methods are more commonly used for unstructured data like images, text, and speech \cite{reviewOfImbalanced}.

It was observed in the related papers that filter and wrapper feature selection methods were the most frequently utilised. Regarding filter methods, a variety of metrics were employed to rank the features, while heuristic search was a common choice for wrapper methods. It is also found the frequent use of feature selection and extraction for addressing real-world problems such as disease diagnosis, textual sentiment analysis, fraud detection, and other rare event detection problems  \cite{reviewOfImbalanced}.

\subsection{Cost-sensitive learning}

\begin{figure}
    \centering
    \includegraphics[width=\linewidth]{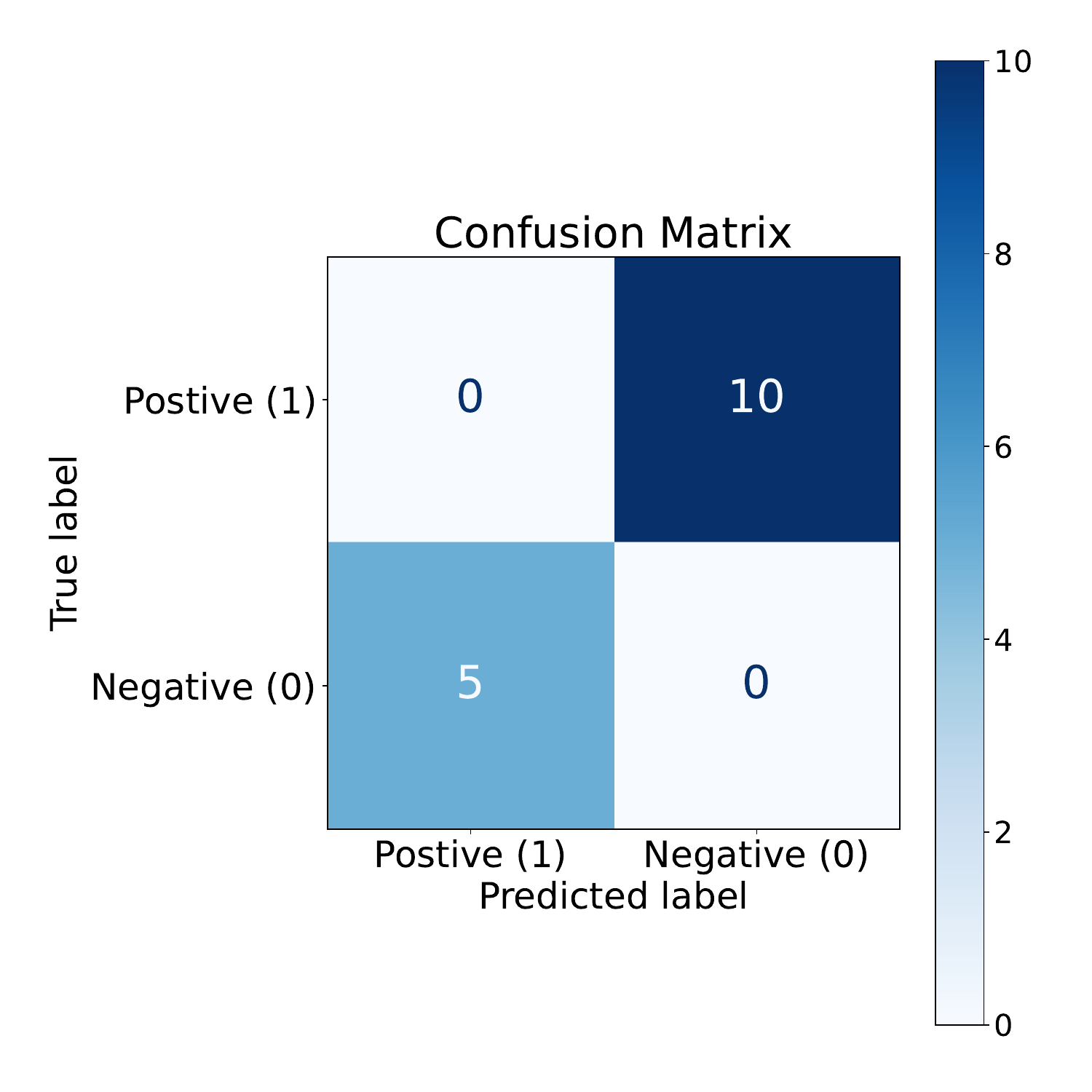}
    \caption{Sample Cost Matrix}
    \label{fig:1}
\end{figure}

Cost-sensitive learning considers the varying costs of misclassifying different classes. Cost matrices are commonly used to define these costs, with $C_{ij}$ indicating the cost of misclassifying examples from class $i$ as class $j$. In specific fields, domain experts can determine cost matrices, resulting in a fixed cost matrix. Alternatively, in data stream scenarios, these matrices may vary at each optimisation loop step based on the algorithm's efficiency, which is referred to as an adaptive cost matrix\cite{ghazikhani2013, reviewOfImbalanced, Lopez2013}. For example, a fixed cost matrix with binary classes \{0, 1\}, corresponding to the first condition, is depicted in Fig. \ref{fig:1}. The cost of misclassifying a sample to class 0 is 10, while the cost of misclassifying to class 1 is 5. The reason for the higher cost of misclassifying to class 0 is that our primary interest lies in the minority class, which we have defined as the positive class (1).

By assuming higher costs for incorrectly classifying samples from the minority class compared to the majority class, cost-sensitive learning can be integrated at both the data level (such as re-sampling and feature selection) and the algorithmic level. At the algorithmic level, the primary concept is to establish direct cost-sensitive learning, which involves incorporating and utilising misclassification costs into the learning algorithms. Regarding the data level, cost-sensitive learning concerns the incorporation of a `preprocessing' mechanism for the training data or a `postprocessing' of the output in a manner that does not alter the original learning algorithm \cite{Lopez2013}. This category can be further divided into two primary categories: thresholding and sampling:

\begin{itemize}
    \item \textbf{Thresholding} operates on the principle of basic decision theory, which involves assigning instances to the class with the lowest expected cost. For example, in a binary classification problem, a typical decision tree assigns the class label of a leaf node based on the majority class of the training samples that reach the node. A cost-sensitive algorithm designates the class label to the node that minimizes the classification cost.
    \item \textbf{Sampling} is associated with modifying the training dataset. The most common technique is to adjust the original class distribution of the training dataset based on the cost decision matrix through under-sampling/over-sampling or assigning instance weights.
\end{itemize}

Guo Haixiang et al. \cite{reviewOfImbalanced} provided a comprehensive overview of the cost-sensitive learning methods published in the last decade in Table \ref{tab:1}. They also concluded that in comparison to resampling methods, cost-sensitive learning is more computationally efficient and may be better suited for handling big data streams. However, they noted that this approach is significantly less popular than resampling methods. One potential explanation is that resampling is a more common choice for researchers not well-versed in machine learning. Unlike cost-sensitive learning, which often requires modification of the learning algorithm, resampling methods are much simpler to implement directly in both single and ensemble models.

\begin{table*}[h!]
\caption{Summary of cost-sensitive learning methods}
\label{tab:1}
\resizebox{\textwidth}{!}{%
\begin{tabular}{ll}
\hline
\multicolumn{1}{c}{\textbf{Method}} & \multicolumn{1}{c}{\textbf{Detail Strategy}} \\ \hline
\begin{tabular}[c]{@{}l@{}}Methods based on training data \\ modification\end{tabular} & \begin{tabular}[c]{@{}l@{}}Modifying the decision thresholds or assigning weights \\ to instance when resampling the training dataset according \\ to the cost decision matrix\end{tabular} \\
\hline
\multirow{5}{*}{\begin{tabular}[c]{@{}l@{}}Changing the learning process or learning \\ objective to build a cost-sensitive classifier\end{tabular}} & \begin{tabular}[c]{@{}l@{}}Modifying the objective function of SVM/ELM using a\\ weighting strategy\end{tabular} \\
\cline{2-2}
 & \begin{tabular}[c]{@{}l@{}}Tree-building strategies that could minimise\\ misclassification costs\end{tabular} \\ \cline{2-2}
 & \begin{tabular}[c]{@{}l@{}}Integrating a cost factor into the fuzzy rule-based\\ classification system\end{tabular} \\
 \cline{2-2}
 & Cost sensitive error function on neural network \\
 \cline{2-2}
 & Cost-sensitive boosting methods \\ \hline
Methods based on Bayes decision theory & \begin{tabular}[c]{@{}l@{}}Incorporating cost matrix into Bayes based decision\\ boundary\end{tabular} \\
\hline
\end{tabular}%
}
\end{table*}

\subsection{Ensemble methods}

Ensemble-based classifiers, also known as multiple classifier systems, are recognised for enhancing the performance of a single classifier by combining multiple base classifiers that individually outperform it. Ensembles of classifiers have emerged as a potential solution to address the imbalanced data
classification problem. Ensemble-based methods involve a blend of ensemble learning algorithms and one of the previously discussed techniques, namely data preprocessing ensembles or cost-sensitive learning solutions. When a data-level approach is added to the ensemble learning algorithm, the new hybrid method typically preprocesses the data before training each classifier. On the contrary, cost-sensitive ensembles guide the cost minimisation procedure through the ensemble learning algorithm instead of modifying the base classifier to accommodate costs during the learning process.

\begin{figure*}
    \centering
    \includegraphics[width=\textwidth]{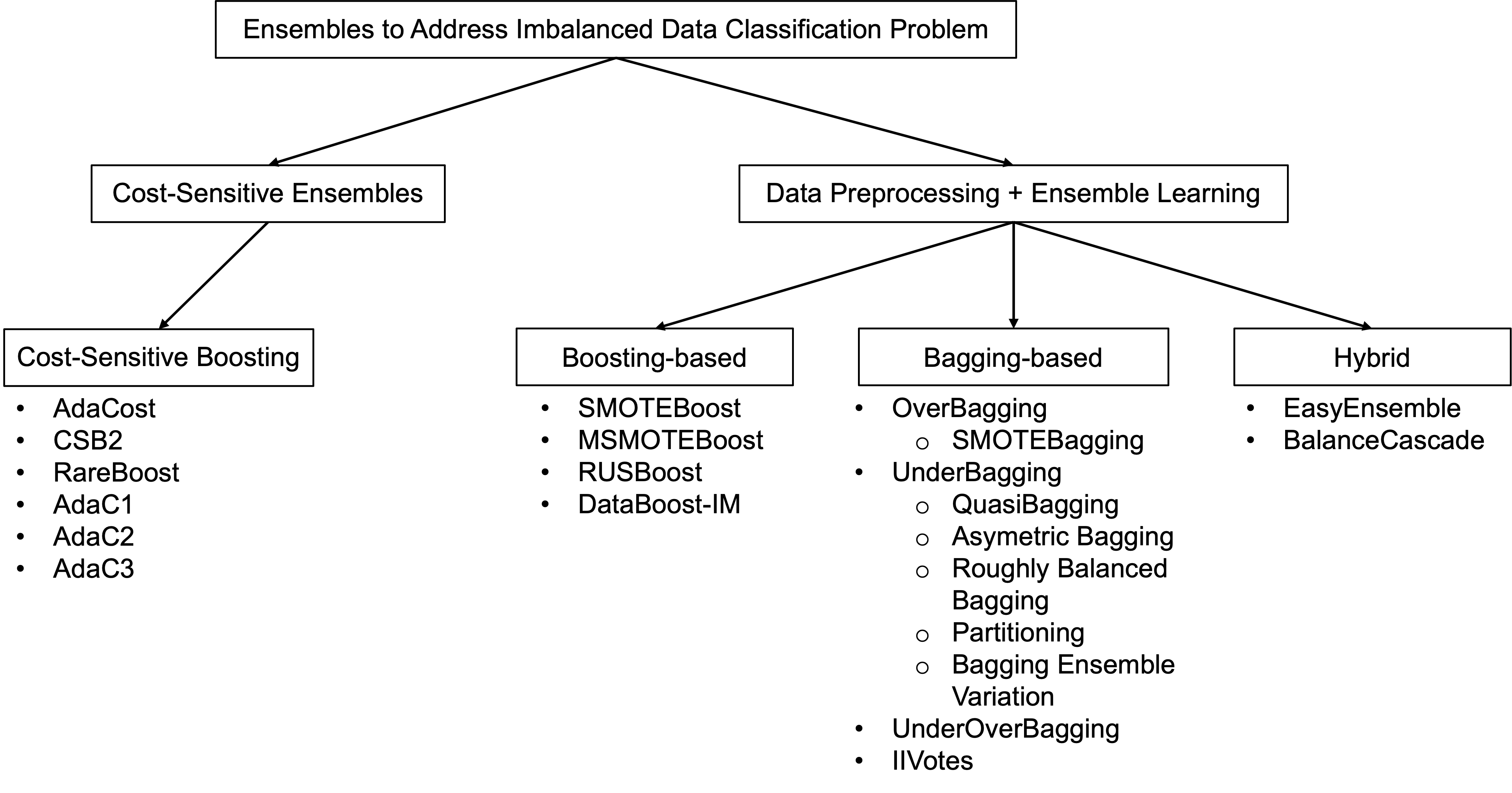}
    \caption{Galar et al.’s proposed taxonomy for ensembles to address the imbalanced data classification problem}
    \label{fig:2}
\end{figure*}

A comprehensive taxonomy for ensemble methods for learning with imbalanced classes can be referenced in \cite{galar2011}, as summarised in Fig. \ref{fig:2}. The authors categorise four distinct families among ensemble approaches for imbalanced learning. They identify cost-sensitive boosting approaches, which are similar to cost-sensitive methods, but with the cost minimisation procedure guided by a boosting algorithm. Additionally, they distinguish three more families sharing a common feature: all embed a data preprocessing technique in an ensemble learning algorithm. These three families are categorised based on the ensemble learning algorithm used, i.e. boosting, bagging, and hybrid ensembles. According to the study, the authors concluded that ensemble-based algorithms are worthwhile because they surpass the performance of solely employing preprocessing techniques prior to training the classifier.

\section{Preliminary}
Given the input data $\mathbf{X}=\left\{\mathbf{x}_1, \mathbf{x}_2, \ldots, \mathbf{x}_n\right\}$, where each $\mathbf{x}_i \in \mathbb{R}^d$ is a feature vector representing the state of the market at time $i$. The feature vector may include various indicators, such as Price changes $(\Delta P)$, Volume traded $(V)$, Bid-ask spread $(S)$, Order book imbalance $(O B I)$, Volatility $(\sigma)$.

Each instance $\mathbf{x}_i$ has a corresponding label $y_i \in\{1,2, \ldots, K\}$, where $K$ is the number of possible outcomes or classes. For example, $K=2$ might represent a binary classification task, such as predicting whether the price will go up or down in the next time interval. In particular, we consider the three-class classification problem in the horizon 1 min, where 
\[
y_i =
\begin{cases} 
1 & \text{if } R_i > \text{fee} \\
-1 & \text{if } R_i < -\text{fee} \\
0 & \text{if } |R_i| \leq \text{fee}
\end{cases}
\]

where:
\begin{itemize}
    \item \( R_i \) is the forward 1 minute return for instance \( i \).
    \item \text{fee} is the trading fee.
\end{itemize}

\textbf{Objective.}
The objective is to learn a classification function $f: \mathbb{R}^d \rightarrow\{1,2, \ldots, K\}$ that maps each input vector $\mathbf{x}_i$ to its corresponding label $y_i$. This function $f$ is often parameterized by a model $\theta$, such as a logistic regression, neural network, or other machine learning model.

Note that in the short horizon such as 1 min, most of the return can not cover the trading fee, which results in the case where most of the label $y_i$ takes value of 0, i.e., highly unbalanced label.

\section{Methodology}
\subsection{Overview}
\begin{figure*}
    \centering 
    \includegraphics[width=0.8\textwidth]{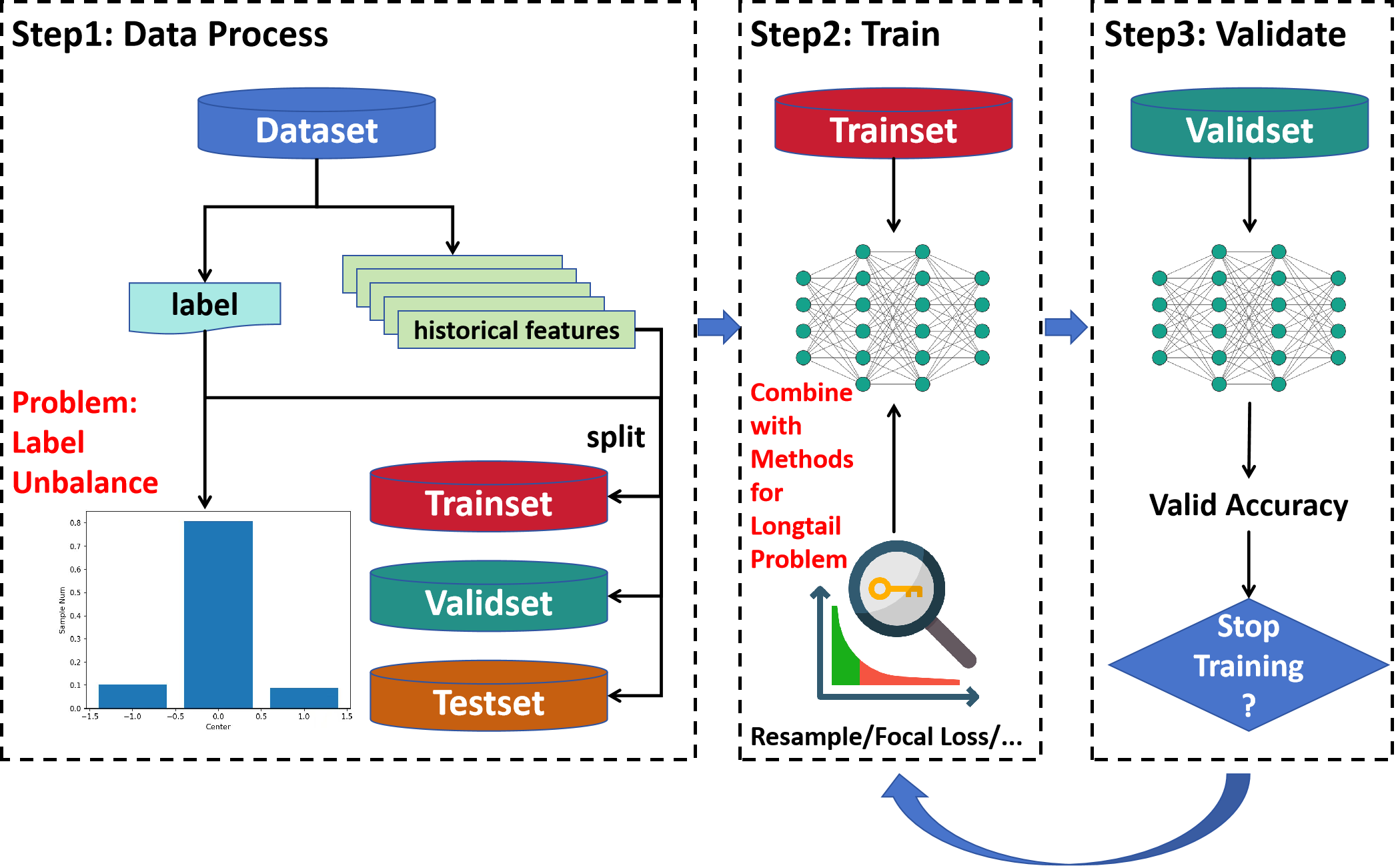}
    \caption{Workflow of Proposed Method}
    \label{fig:main}
\end{figure*}

As shown in Fig. \ref{fig:main}, our method consists of three main phases during training: data processing, training, and validation.

\textbf{(1) Data Processing:} Details regarding the data structure of our dataset and the data processing methods are outlined in Section \ref{sec:Dataset Description}. In summary, we represent the data features as a 13-dimensional vector for each second and utilize the preceding 60 seconds of data to predict the current return, which is categorized into three classes: -1, 0, and +1. However, we identified a significant label imbalance issue in the dataset, with approximately 80\% of samples belonging to class 0, while classes +1 and -1 each account for only about 10\%. To address this, we implemented strategies for tackling long-tail distribution challenges during the training phase. The dataset is then divided into training, validation, and test sets in a 8:1:1 ratio. To prevent information leakage, we split the dataset chronologically rather than through random sampling. Lastly, we offer an optional normalization operation for each sample:
\begin{equation}
\begin{aligned}
& \mu^{(j)}=\frac{\sum_{i=1}^{60} x_i^{(j)}}{60} \ ,\\
& \sigma^{(j)}=\sqrt{\frac{\sum_{i=1}^{60} (x_i^{(j)}-\mu^{(j)})^2}{60}} \ ,\\
& \text{Norm}(x_i^{(j)})=\frac{x_i^{(j)}-\mu^{(j)}}{\sigma^{(j)}} \ ,
\label{norm}
\end{aligned}
\end{equation}
where $x_i^{(j)}$ represents the $j^{th}$ dimension of the $i^{th}$ position in sample $x$, $\mu$ and $\sigma$ represent the mean and standard deviation, respectively. Although this normalization has proven effective in mitigating covariate shift and enhancing model generalization across various fields \cite{covariate_shift, csi-bert}, we observe significant variance in our experiments. Its impact appears to differ across neural networks and methods for addressing label imbalance. At this stage, we cannot definitively categorize its effect as beneficial or detrimental in our task, warranting further investigation.

\textbf{(2) Training:} During this phase, we utilize the training set to train a neural network, incorporating strategies to mitigate label imbalance. We will elaborate on our network architecture and approaches to address label imbalance later in this section.

\textbf{(3) Validation:} To determine model convergence, we employ a standard early stopping mechanism. We monitor the model's accuracy at the end of each epoch and cease training when the accuracy does not improve over several consecutive epochs.

\subsection{Backbone Models}
\subsubsection{Multilayer Perceptron (MLP)}
\begin{figure}
    \centering
    \includegraphics[width=\linewidth]{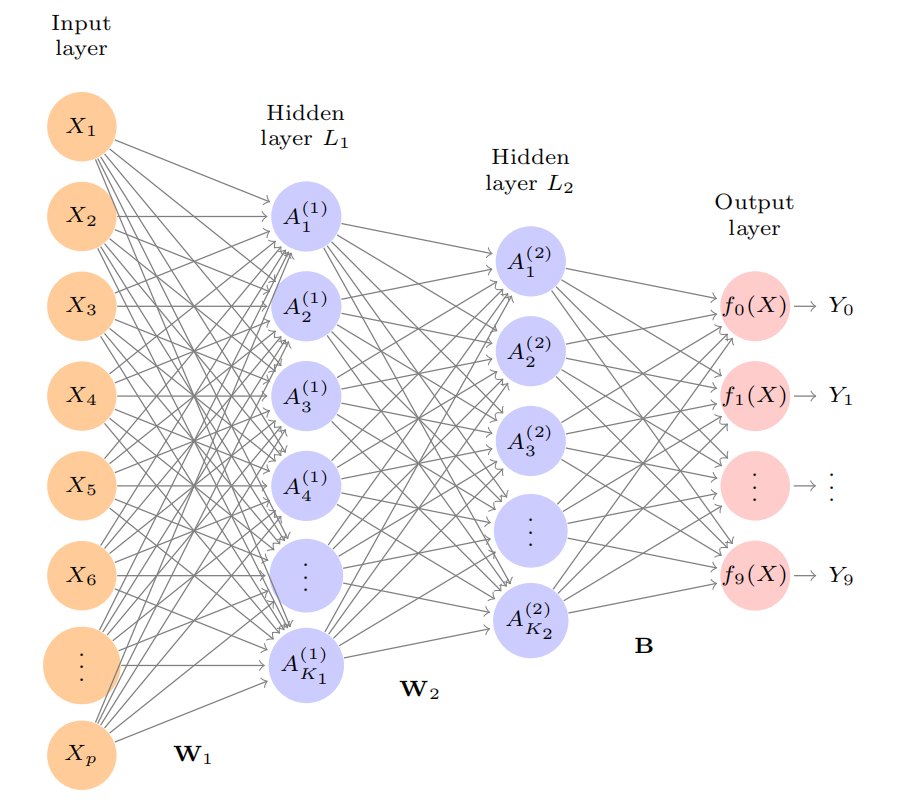}
    \caption{Enter Caption}
    \label{fig:4}
\end{figure}

A Multilayer Perceptron (MLP) is a type of neural network that consists of multiple layers, including an input layer, hidden layers, and an output layer, where each layer is made up of perception nodes (neurons). The nodes in the hidden layers are referred to as activations. Fig. \ref{fig:4} shows an MLP featuring two hidden layers, along with an input and an output layer. An activation $j$ in layer $d+1$ can be formulated as:

\begin{equation}
\begin{aligned}
    A_j^{(d+1)} &= h_j^{(d+1)}(\boldsymbol{X})\\
    &= g(w_{j0}^{(d+1)} + \sum_i w_{ji}^{(d+1)}A_i^{(d)})
\end{aligned} 
\end{equation}
where $g(z)$ is a nonlinear activation function specified beforehand, $\boldsymbol{X}$ is the input vector of $p$ variables, and the superscript notation indicates to which layer the neuron and weights (coefficients) belong. Since each activation is directly a function of the activations $A_i^{(d)}$ from the previous layer $d$ down to the input layer, they are inherently a function of the input vector $\boldsymbol{X}$. Through successive transformations, the network can create fairly intricate transformations of $\boldsymbol{X}$ that ultimately serve as features fed into the output layer.

After the transformations, we proceed to the output layer. Let $Z_m$ denote the $m$th possible output, which is computed similarly to the activations. If the output $Y$ is continuous, we simply set $\hat Y = Z_m$. For categorical responses, however, we apply activation functions such as the sigmoid function to produce the output $\hat Y_m$ \cite{ISLP}.

In our implementation, both the activation functions in the hidden layers and the output layer utilise the LeakyReLU function, defined as follows:

\begin{equation}
\text{LeakyReLU}(x) =
    \begin{cases}
        x & \text{if $x \geq 0$}\\
        \text{negative\_slope}\times x, & \text{otherwise}
    \end{cases}
\end{equation}
where negative\_slope is a hyperparameter.

\subsubsection{Long Short-Term Memory (LSTM)}
Recurrent Neural Networks (RNNs) are specifically created to handle and leverage sequential input data, such as financial time series data. 
Let $X=\{X_1,X_2,\dots,X_T\}$, where each vector $X_t$ in the input sequence contains $p$ components and the hidden layer has $K$ units. The activation can be expressed as:
\begin{equation}
    A_{tk}=g(w_{k0}+\sum_{j=1}^p w_{kj}X_{tj}+\sum_{s=1}^K u_{ks}A_{t-1,s})
\end{equation}
where $w_{kj}$ are weights for input layer, and $u_{ks}$ are the weights for the hidden-to-hidden layers. The resulting output is given by:
\begin{equation}
    O_t = \beta_0 + \sum_{k=1}^K\beta_kA_{tk}
\end{equation}
where weights $\beta_k$ are the weights corresponding to the output layer, used for a quantitative prediction or with an additional sigmoid function for a binary outcome. The weights $w_{kj}$, $u_{ks}$ and $\beta_k$ are consistently applied as each element of the sequence is processed, exemplifying the weight sharing characteristic inherent to RNNs \cite{ISLP}.

\begin{figure}
    \centering
    \includegraphics[width=\linewidth]{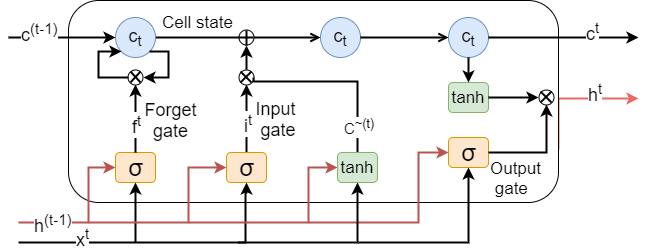}
    \caption{Long Short Term Memory (LSTM) cell}
    \label{fig:lstm}
\end{figure}

Conventional RNNs have difficulty learning tasks that require long-term dependencies because of issues with vanishing gradients and the decay of error flow. LSTMs provide a novel architecture that preserves a consistent error flow and tackles the vanishing gradient issue by using specialized units (memory cells) that retain their internal state over time, enabling them to "remember" information for extended periods. From Figure \ref{fig:lstm}, we can see the layout of a memory cell. Three main components oversee the functioning of the memory cells \cite{LSTM}:
\begin{enumerate}
    \item Input Gate: Determines when new input is introduced to the memory cell, denoted by $i$.
    \item Forget Gate: Decides when to erase or reset the cell’s state, denoted by $f$.
    \item Output Gate: Controls when the state of the memory cell is sent out to the rest of the network, denoted by $o$.
\end{enumerate}

If we denote the collection of weights $w_{kj}$ and $u_{ks}$ as $\bold W$ and $\bold U$ respectively, we can express the three components as follows:

\begin{equation}
    \begin{aligned}
        i_t &= \sigma(\bold W x_t + \bold U h_{t-1}) \\
        f_t &= \sigma(\bold W x_t + \bold U h_{t-1}) \\
        o_t &= \sigma(\bold W x_t + \bold U h_{t-1})
    \end{aligned}
\end{equation}
where $\sigma()$ is the sigmoid function, $x_t$ is the input vector at sequence $t$, and $h_{t-1}$ is the hidden state at sequence $t-1$. The relationship between the cell's state $c_t$ and the three gates is

\begin{equation}
    \begin{aligned}
        c_t &= f_t \odot c_{t-1} + i_t \odot  \tanh(\bold W x_t + \bold U h_{t-1}) \\
        h_t &= o_t \odot \tanh(c_t)
    \end{aligned}
\end{equation}

\subsubsection{Mamba}
Time series processing has long been a prominent topic in deep learning. Following the development of RNN \cite{RNN}, GRU \cite{GRU}, and LSTM \cite{LSTM}, Transformer \cite{Transformer} has garnered significant attention for time series tasks due to its attention mechanism, which effectively captures relationships between any two positions in a sequence. However, there is still considerable debate regarding its effectiveness. For instance, some researchers argue that the Transformer struggles to recognize order due to its positional invariance \cite{transformer-time}.

Recently, Mamba \cite{Mamba}, developed on structured State Space Sequence Models (SSMs) \cite{gu2021efficiently,gu2021combining}, has gained widespread attention and is seen as a potential replacement for the Transformer. In traditional SSMs, the workflow can be expressed as:
\begin{equation}
\begin{aligned}
h'(t) & = Ah(t) + Bx(t) \\
y(t) & = Ch(t)
\end{aligned} 
\end{equation}
where $A,B,C$ are model parameters, $x(t), y(t)$ represent the model's input and output, $h(t), h'(t)$ denote the hidden state before and after updating, and $t$ indicates time. Mamba enhances this process through a selection mechanism, allowing $A,B,C$ to depend on the input by utilizing a neural network, rather than being fixed matrices. This approach is quite similar to the attention mechanism in Transformers: for different tokens, Mamba employs distinct processing patterns for the hidden state, which incorporates historical information. Compared to Transformer, Mamba offers faster inference speeds and demonstrates excellent performance across various sequence-level tasks, such as audio processing and NLP.


\subsection{Under-sampling}
The poor performance of models trained directly on the full dataset arises from an unbalanced data distribution, which causes the model to focus more on the class with a larger number of samples. Therefore, our initial approach involves the implementation of under-sampling techniques to achieve equilibrium among the labels. In our dataset, the proportions of classes -1, 0, and +1 are approximately 1:8:1. Consequently, in each epoch, we randomly drop $\frac{1}{8}$ of the samples from class 0 in each epoch.

\subsection{Cost-sensitive Learning}
\subsubsection{Fixed Cost Matrix}
In the context of addressing label imbalance, cost-sensitive learning at the algorithmic level aims to focus the model's attention on classes with fewer samples, similar to the intuition behind resampling methods. Initially, we employed a straightforward method, direct loss weighting, to tackle the label imbalance issue. The loss for a sample can be expressed as:
\begin{equation}
\begin{aligned}
L = \sum_{c=1}^C w_c l_c
\label{eq:loss-weighting}
\end{aligned}
\end{equation}
where $C$ is number of classes, $l_c$ is the loss function for samples from class $c$, and $w_c$ is the assigned weight. For our specific task, we designated the weights for classes -1, 0, and 1 as 8.0, 1.0, and 8.0, respectively. Intuitively, this method serves a similar process to resampling.

Moreover, we adjusted the weights based on the specific size of each class. We incorporated cost into the mean square error (MSE) for each class, building on Castro and Braga's research on imbalanced binary labels \cite{castro2013novel}. The loss function for each sample can be defined as:

\begin{equation}
    L = \sum_{c=1}^C\frac{N_{-c}}{(C-1)\cdot N}(1-p_c)^2 l_c
\label{eq:sensitive-loss}
\end{equation}
where $N_{-c}$ is the number of samples other than class $c$, and $p_c$ is the probability that the sample is in class $c$. The cost is normalised by $C-1$ to ensure that $\sum_{c=1}^C \frac{N_{-c}}{(C-1)\cdot N}=1$. Although fixed, the cost is aligned with the true distribution of the labels rather than being assigned intuitively.

\subsubsection{Adaptive Cost Matrix}
Focal loss \cite{lin2017focal} is a method akin to loss weighting, but it incorporates a dynamic cost matrix for each sample. It can be represented as:
\begin{equation}
\begin{aligned}
FocalLoss(\boldsymbol{p}, y) = - \sum_{c=1}^C (1-p_c)^\lambda log(p_c) \cdot\textbf{1}\{y=c\} \\
CrossentropyLoss(\boldsymbol{p}, y) = - \sum_{c=1}^C log(p_c) \cdot\textbf{1}\{y=c\} 
\label{eq:focal-loss}
\end{aligned}
\end{equation}
where $\boldsymbol{p}$ is the vector output of the probabilities for each class by the model, $y$ is the ground truth, $C$ is the number of classes, $\lambda$ is a control parameter, and $\mathbf{1}\{y=c\}$ is the indicator function. Compared to traditional cross-entropy loss, focal loss assigns greater weights to samples for which the model has low confidence $(p_c)$ in the correct class. This adaptive weight $(1-p_c)^\lambda$ is the cost of misclassifying a sample to another class. The approach is particularly relevant in scenarios with imbalanced labels, where the model typically exhibits lower confidence for classes with fewer samples. Moreover, the dynamic nature of the costs during training ensures they remain relevant throughout the process.

What's more, \cite{zhao2024adversarial} also propose a similar loss function:
\begin{equation}
\begin{aligned}
w_c & = \frac{\frac{1}{a_c}}{\sum_{c=1}^C \frac{1}{a_c}}  \\
L & = \sum_{c=1}^C w_c l_c
\end{aligned}
\end{equation}
where $a_c$ is the average accuracy of the $c^{th}$ class.

\section{Experiment}
\subsection{Dataset Description} \label{sec:Dataset Description}

Our data spans from May 4th, 2023, to May 29th, 2023, covering a total of 20 trading days of high-frequency futures data, with a frequency of 0.5 seconds. The dataset includes six varieties: rebar (rb2310), silver (ag2308), fuel oil (fu2309), nickel (ni2306), tin (sn2306), and gold (au2308). Please note that each variety has different trading hours, resulting in varying sample sizes. Silver and gold have larger sample sizes, while rebar and fuel oil have smaller ones.

The raw data contains key information such as trading time, daily prices (opening, closing, highest, and lowest), latest transaction price ($lastPrice$), cumulative transaction amount, cumulative transaction volume, and the price and volume of buy and sell orders at the first to fifth trading positions ($bidPrice_i$ and $askPrice_i$, where $i=1,2,3,4,5$, which represents 5 positions respectively). Based on this data, we constructed 13 variables to predict returns. These variables include:
\begin{itemize}
    \item $midPrice$, which is the average of the best buy (buy one) and sell (sell one) prices at each data point.
    \item $\mathit{diffBidPrice}_i$ and $\mathit{diffAskPrice}_i, i=1, 2, 3, 4, 5$, which are the price differences between each level and the average $midPrice$ at each data point.
    \item $\mathit{diffLastPrice}$, which is the price difference between the latest transaction price $lastPrice$ and the average $midPrice$.
    \item $logVolume$, which is the logarithm of transaction volume $volume$ in the past 0.5 seconds (if there is no transaction, it is recorded as 0).
\end{itemize}
The calculation formula is as follows:
\begin{equation}
\begin{aligned}
& \mathit{midPrice} = \frac{\mathit{bidPrice}_{1} + \mathit{askPrice}_{1}}{2},\\
& \mathit{diffBidPrice}_i = \mathit{bidPrice}_i - \mathit{midPrice}, i=1, 2, 3, 4, 5,\\
& \mathit{diffAskPrice}_i = \mathit{askPrice}_i - \mathit{midPrice}, i=1, 2, 3, 4, 5,\\
& \mathit{diffLastPrice} = \mathit{lastPrice} - \mathit{midPrice},\\
&
\mathit{logVolume} =
\begin{cases} 
\mathit{log(volume)} & \text{if } \mathit{volume} > 0 \\
0 & \text{if } \mathit{volume} = 0
\end{cases}
\label{variable}
\end{aligned}
\end{equation}

For calculations of return, we compute the rate of change of the average $midPrice$ at time $t$ relative to time $t$-29.5s (An interval of 59 data points, with 0.5 seconds between each data point).

Additionally, we filled forward missing values and recorded the top 59 returns for each transaction segment (23:00 the previous day, 9:00 am, 10:30 am, and 1:30 pm on the current day) as missing values for each trading day.

As shown in Table \ref{tab: factors}, we also constructed several factors to predict the returns so as to test the quality of our data. The accumulated sum of factors of returns for 8 different factors are shown in Fig. \ref{fig:factor}. The significant and unstable differences in the effects of factors indicate the need for models to further analyze and estimate data in order to achieve better predictive performance.

\subsection{Experiment Setup}

The implementation details of our proposed method are presented in Table \ref{tab: configuration}. For our experiments, we utilized an Intel(R) Xeon(R) Silver 4210R CPU @ 2.40GHz and two NVIDIA RTX 4090 GPUs as the hardware devices. The detail of each neural network is shown as Table \ref{tab: networks}.

\begin{table*}
\caption{Factor Construction}
    \centering
        \begin{tabular}{|c|c|c|}
        \hline
        \textbf{Factor Name} & \textbf{Factor Description} \\
        \hline 
        \ mid\_price\_mean & mean of the mid price in the past 30 seconds  \\
        \hline 
        \ mid\_price\_std & standard deviation of the mid price in the past 30 seconds  \\
        \hline 
        \ mid\_price\_skew & skewness of the mid price in the past 30 seconds \\
        \hline 
        \ mid\_price\_kurt & kurtosis of the mid price in the past 30 seconds \\
        \hline 
        \ volume\_pct & the total transaction amount in the past 30 seconds divided by the total transaction amount in the past 5 minutes \\
        \hline
        \ prop\_quoted\_spread & (lowest ask price - highest bid price) / mid price \\
        \hline
        \ beta & the slope of the regression between return and transaction volume \\
        \hline
        \ illiquidity & absolute value of yield divided by transaction volume \\
        \hline
        \end{tabular}
\label{tab: factors}
\end{table*}

\begin{table}
\caption{Experiment Configurations}
    \centering
        \begin{tabular}{|c|c|}
        \hline
        \textbf{Configuration} & \textbf{Our Setting} \\
        \hline 
        Sample Dimension     & $60 \times 13$   \\
        \hline 
        Class Number     & 3  \\
        \hline 
        Batch Size   & 512  \\
        \hline
        Optimizer    & Adam \\
        \hline 
        Learning Rate    & 0.0001  \\
        \hline 
        Early Stop Epochs   & 10 \\
        \hline 
        \end{tabular}
\label{tab: configuration}
\end{table}

\begin{table}
\caption{Model Configurations}
    \centering
    \begin{adjustbox}{width=0.5\textwidth}
        \begin{tabular}{|c|c|c|c|c|c|}
        \hline
        \textbf{Model} & \textbf{Scale} & \textbf{Layer} & \textbf{Important Parameters} & \textbf{GPU Occupation}\\
        \hline 
        \textbf{MLP}  & 54K & 4 & structure=[780,64,64,3] & 2.82G \\
        \hline 
        \textbf{LSTM \cite{LSTM}}  & 26K & 4 & hidden\_dim=64 & 3.66G \\
        \hline 
        \textbf{Mamba \cite{Mamba}} & 339K & 4 & state\_size=4 & 4.73G \\
        \hline 
        \end{tabular}
    \end{adjustbox}
\label{tab: networks}
\end{table}

\subsection{Experiment Result \& Analysis}

\begin{figure*}
\centering 
\subfloat[Au]{\includegraphics[width=0.3\textwidth]{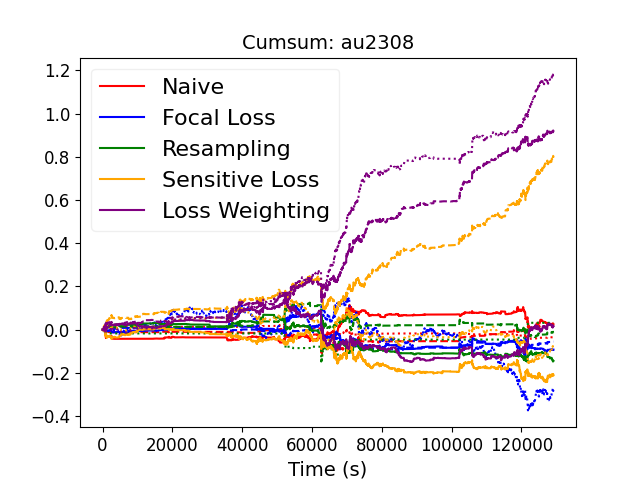}}
\subfloat[Ag]{\includegraphics[width=0.3\textwidth]{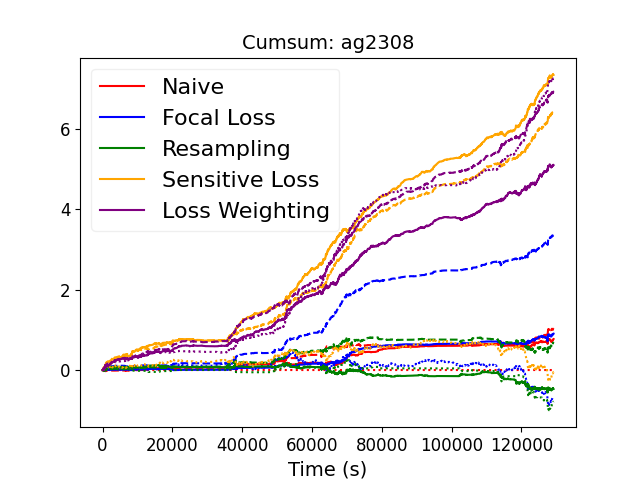}}
\subfloat[Fu]{\includegraphics[width=0.3\textwidth]{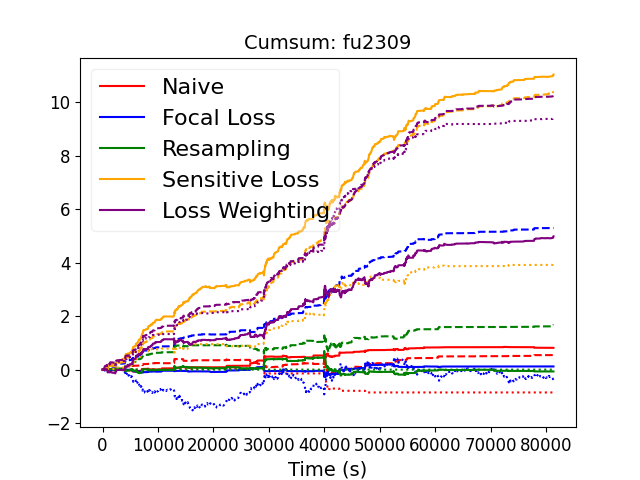}} \\
\subfloat[Ni]{\includegraphics[width=0.3\textwidth]{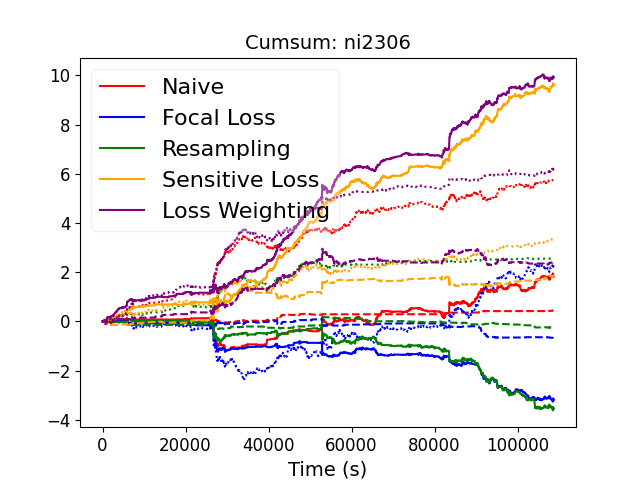}}
\subfloat[Rb]{\includegraphics[width=0.3\textwidth]{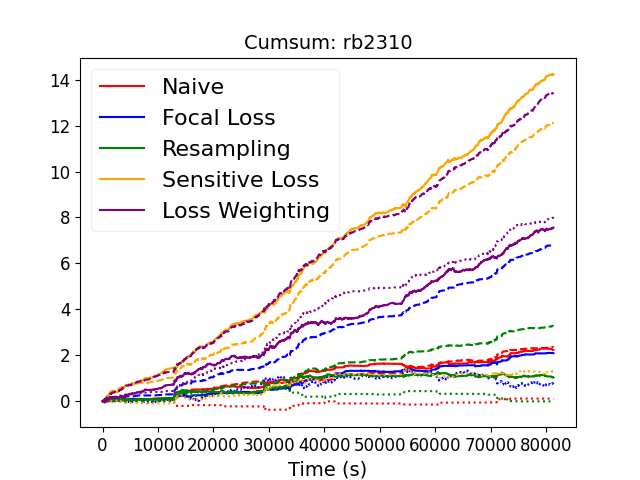}}
\subfloat[Sn]{\includegraphics[width=0.3\textwidth]{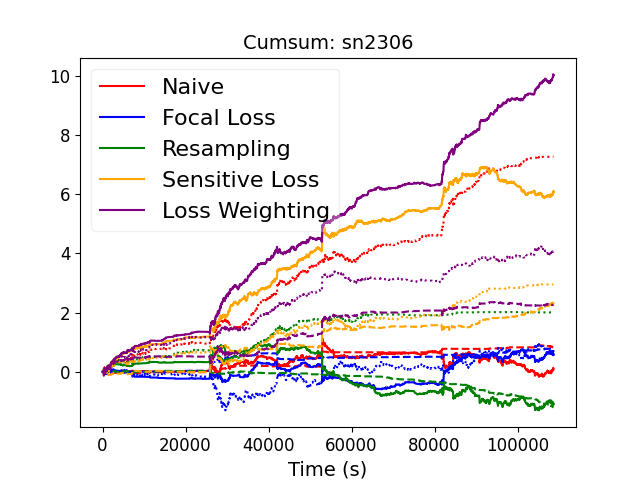}} \\
\caption{Experimental Results for 6 Different Items: Different colors represent various methods for addressing label imbalance, while different line styles indicate different backbone models. The solid line represents Mamba \cite{Mamba}, the dashed line represents LSTM \cite{LSTM}, and the dotted line represents MLP.}
\label{fig:exp}
\end{figure*}

\begin{figure*}
\centering 
\subfloat[mid\_price\_mean]{\includegraphics[width=0.22\textwidth]{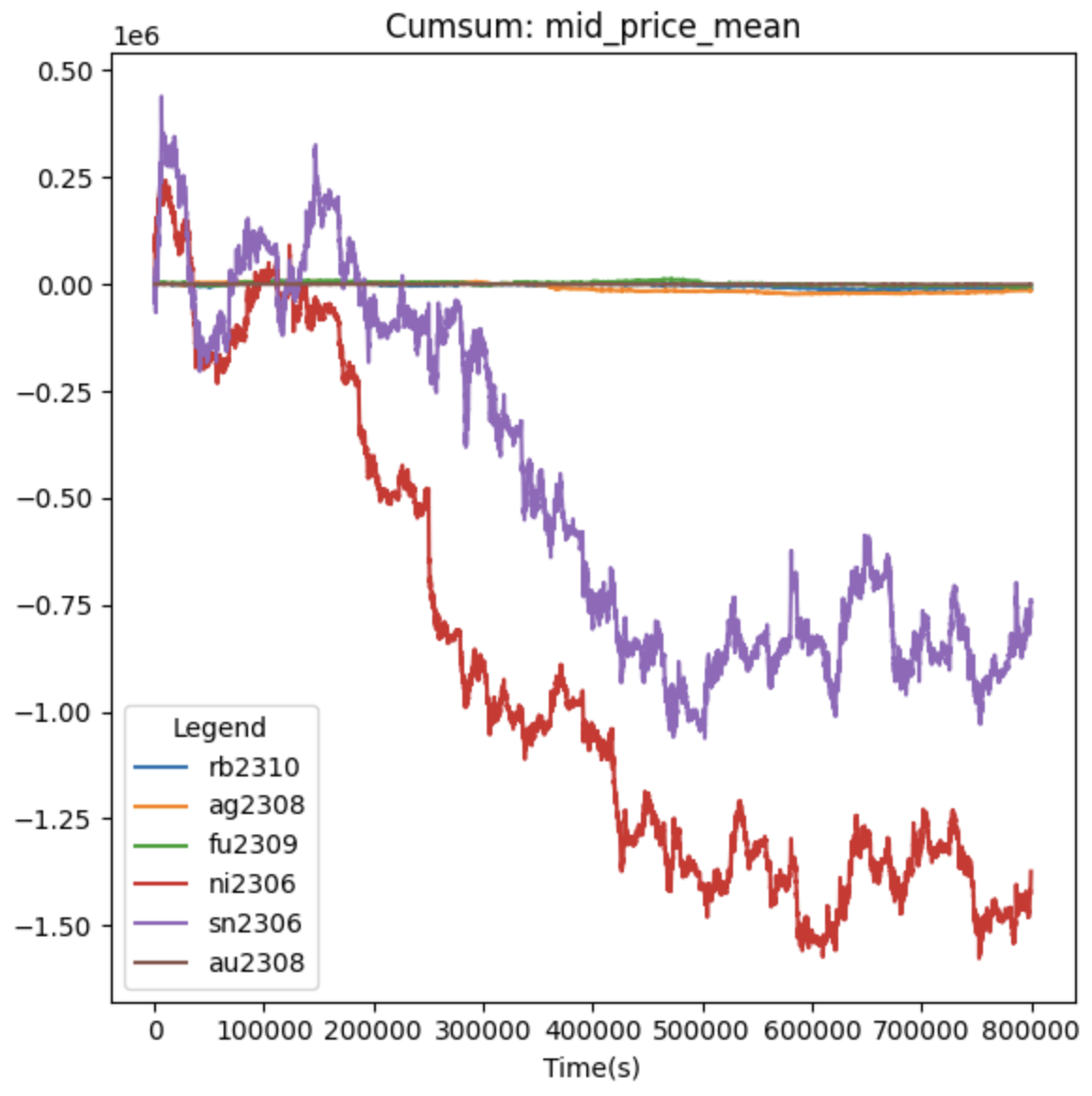}}
\subfloat[mid\_price\_std]{\includegraphics[width=0.22\textwidth]{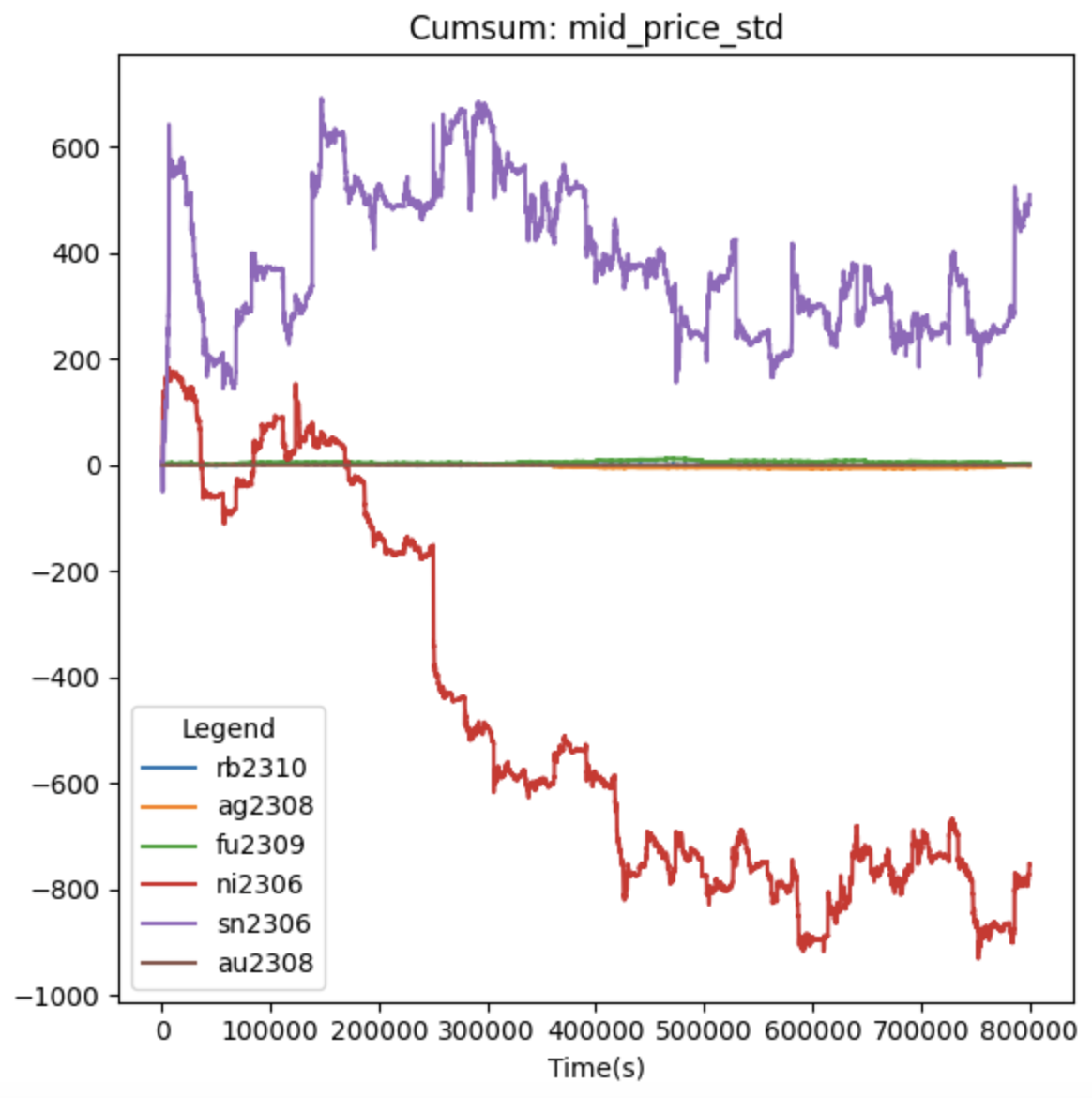}}
\subfloat[mid\_price\_skew]{\includegraphics[width=0.22\textwidth]{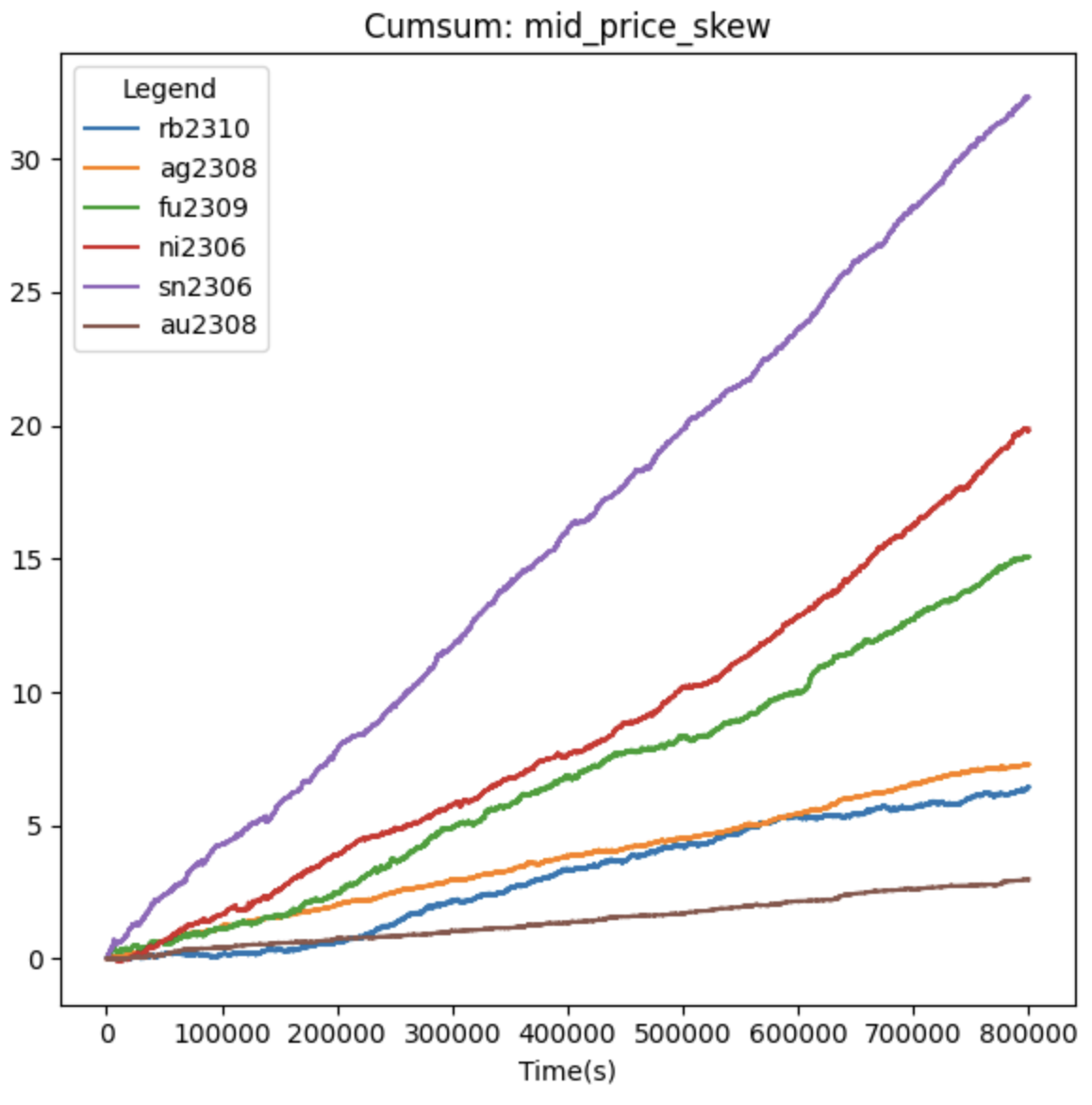}} 
\subfloat[mid\_price\_kurt]{\includegraphics[width=0.22\textwidth]{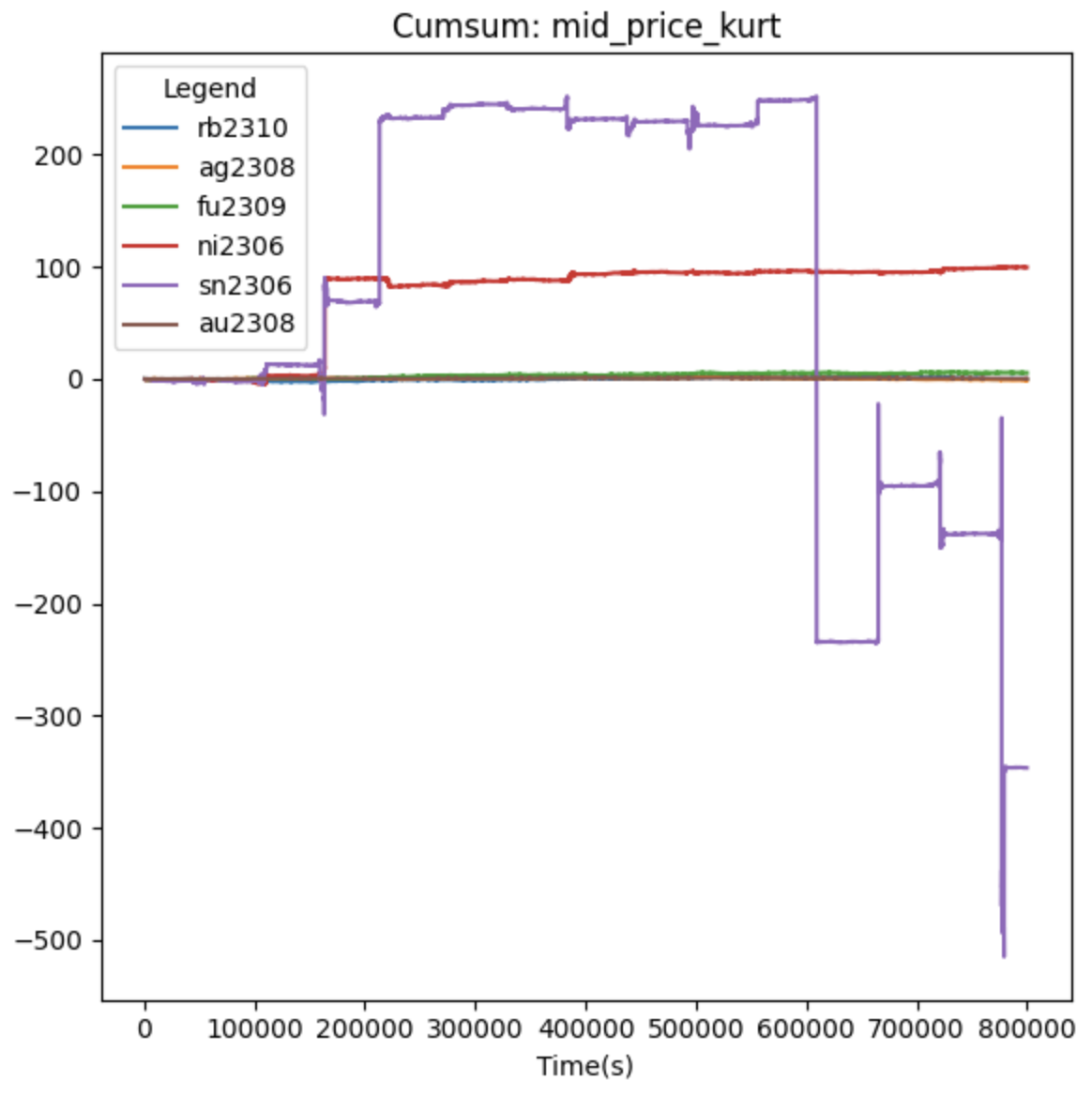}} \\ 
\subfloat[volume\_pct]{\includegraphics[width=0.22\textwidth]{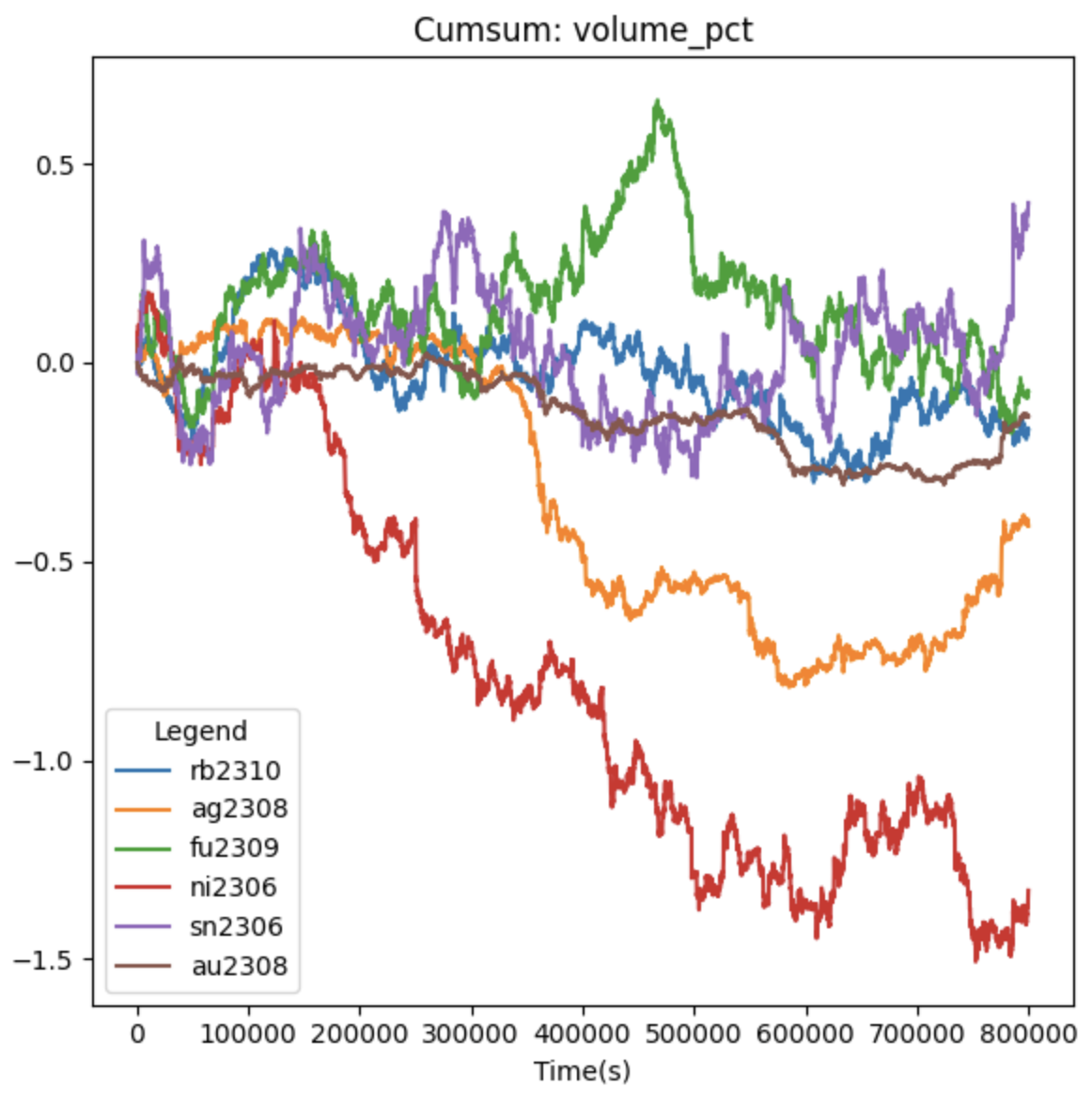}}
\subfloat[prop\_quoted\_spread]{\includegraphics[width=0.22\textwidth]{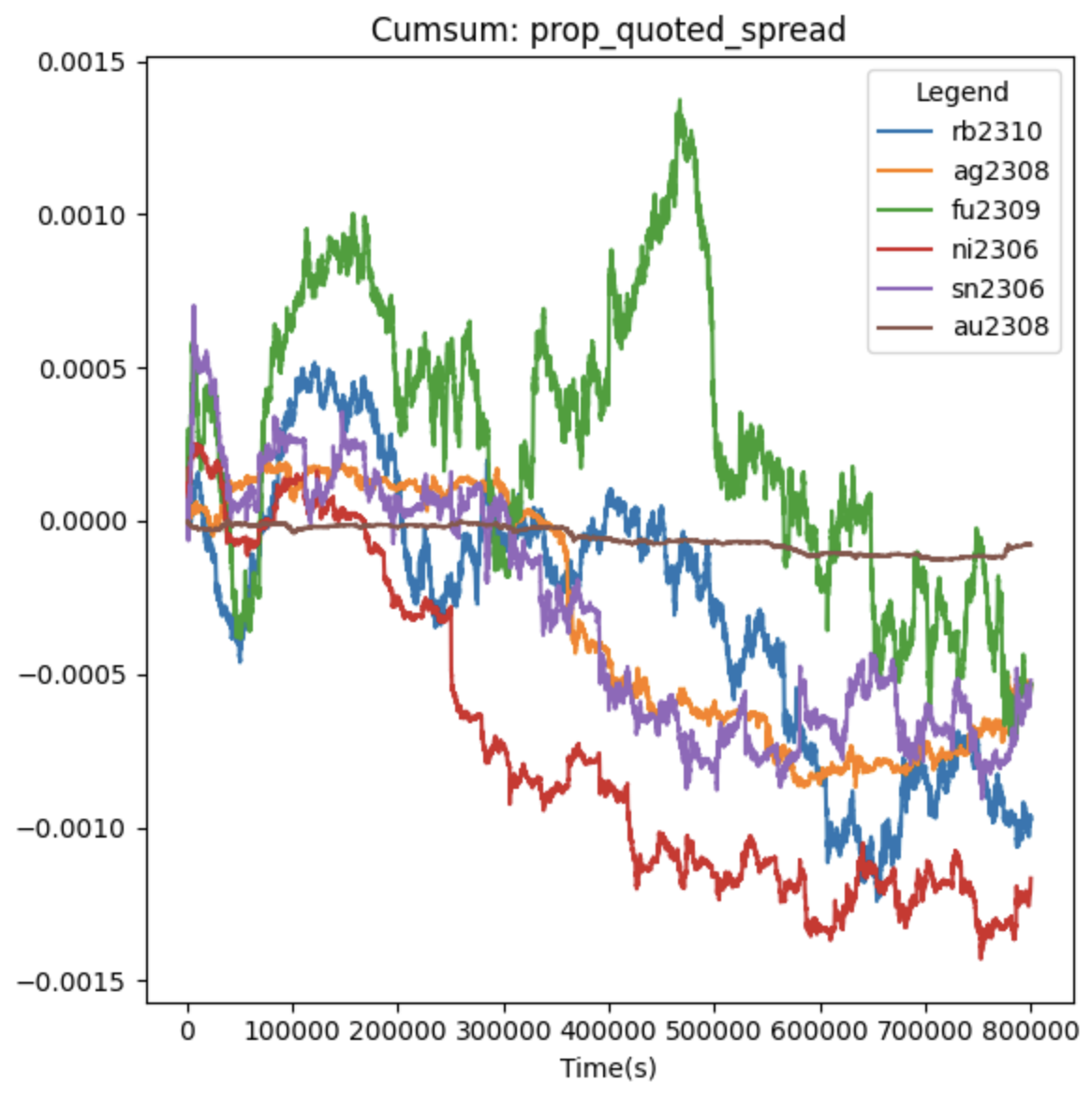}}
\subfloat[beta]{\includegraphics[width=0.22\textwidth]{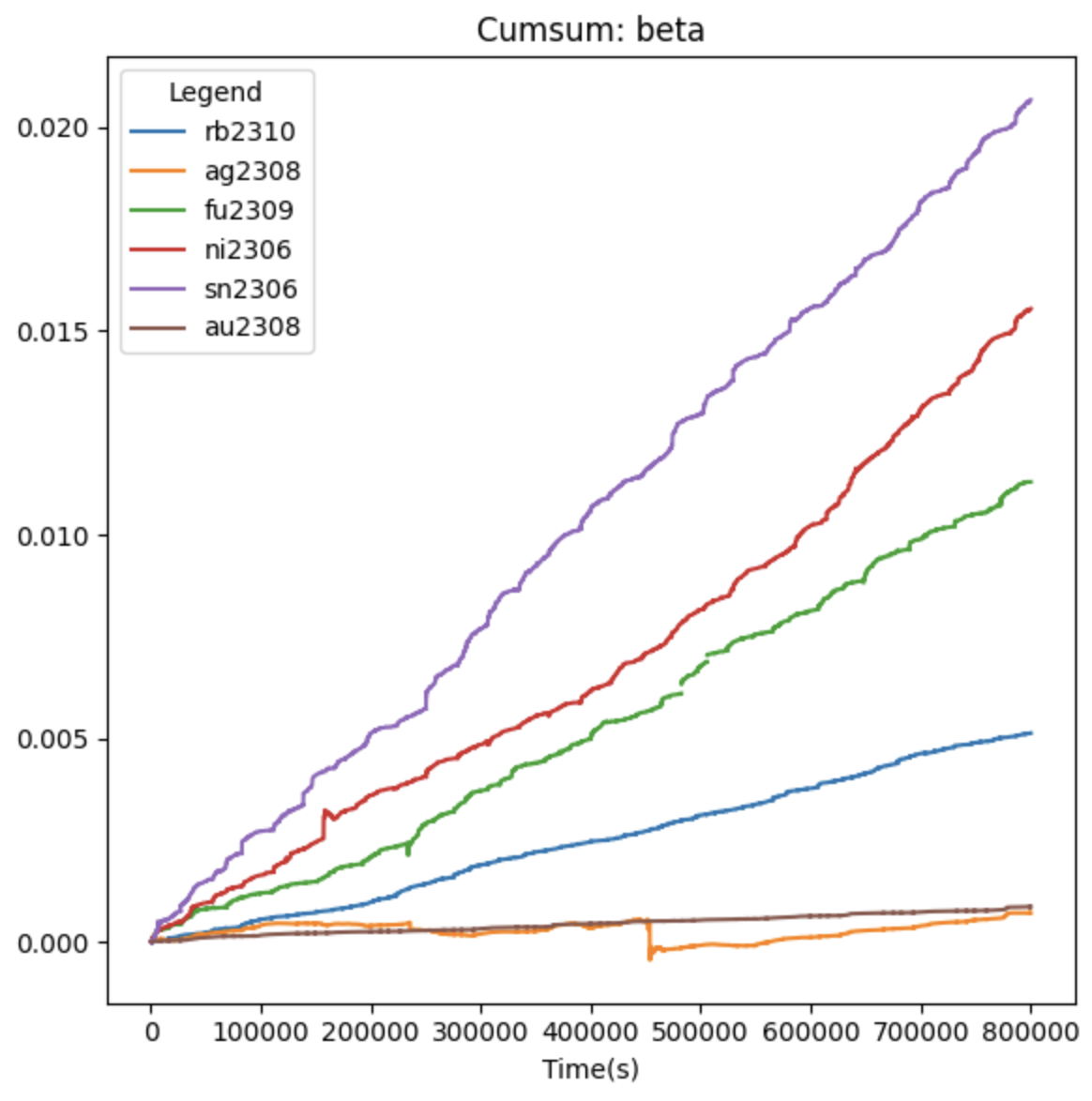}}
\subfloat[illiquidity]{\includegraphics[width=0.22\textwidth]{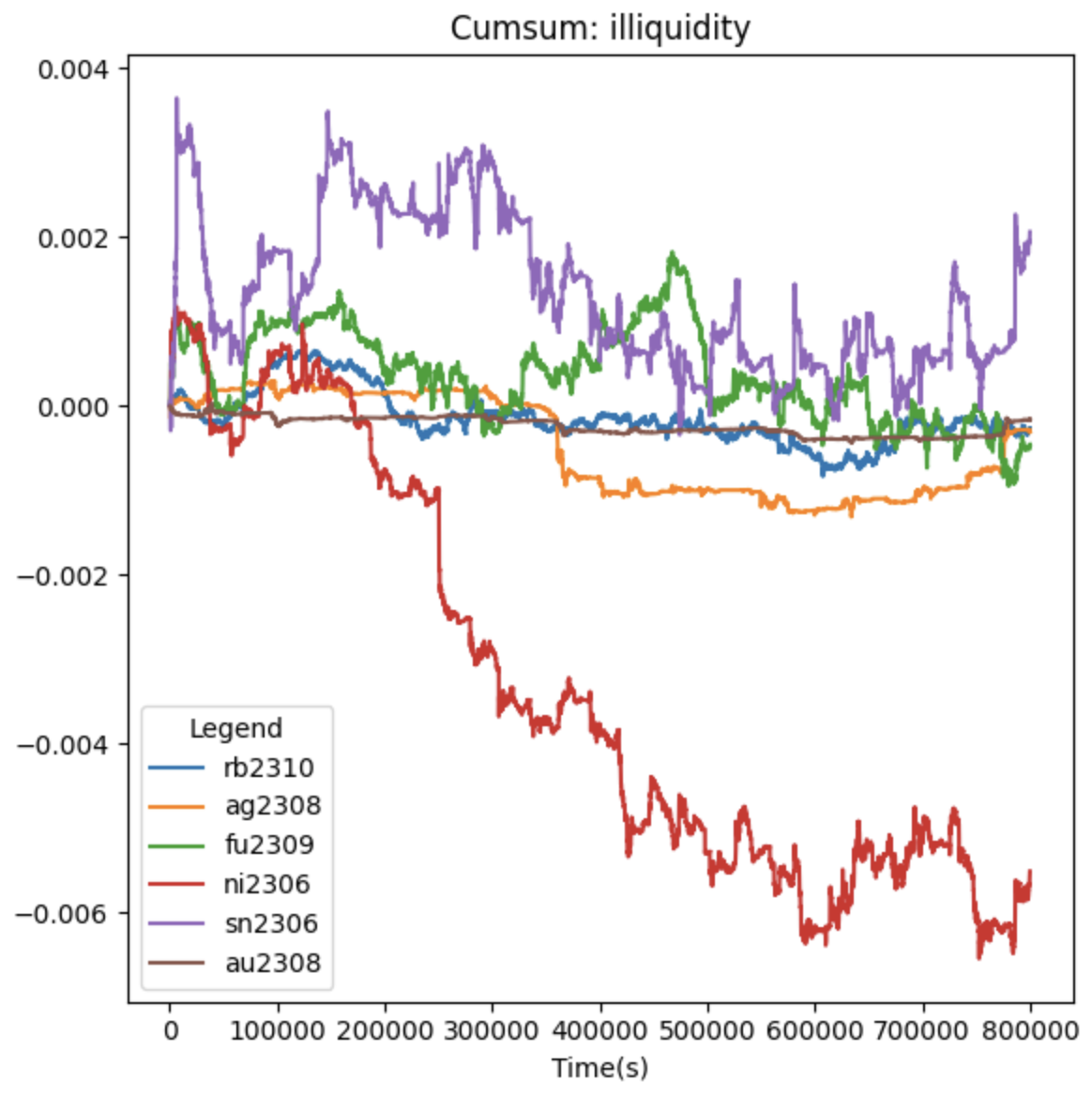}} \\
\caption{Accumulated Sum of Factors Multiplied by Returns for 8 Different Factors: Different colors represent 6 different futures varieties.}
\label{fig:factor}
\end{figure*}

The experimental results are shown in Fig. \ref{fig:exp}. We observed that the model is prone to overfitting when the training set consists of only one item. This may be due to the limited amount of data, which contains a significant amount of noise. Consequently, we utilized all six items to train the models and evaluated their performance across different items.

Regarding the backbone models, it is difficult to determine which one is the best. The LSTM \cite{LSTM} and Mamaba \cite{Mamba} models generally outperform the MLP in most scenarios, likely because they are better suited to capture temporal relationships due to their structural design. However, we noted that the training time for Mamaba is significantly greater than for the others, as it cannot compute in parallel. Although this comparison may be unfair due to its higher number of parameters, further reduction of its scale proves challenging.

For addressing the label imbalance problem, it appears that the sensitive loss (Eq. \ref{eq:sensitive-loss}) and loss weighting (Eq. \ref{eq:loss-weighting}) methods yield better performance than others. In our task, both resampling and focal loss (Eq. \ref{eq:focal-loss}) \cite{lin2017focal} methods sometimes performed worse than the benchmark, which employed no specific approach for the label imbalance problem. The reasons for this require further investigation.

\section{Discussion \& Future Directions}
In this project, we primarily illustrate the efficiency of using machine learning methods in high-frequency trading with label imbalance. In addition to the findings reported in this paper, we aim to highlight some problems and challenges encountered during our project to provide insights and experiences for future research.

\subsection{Data Noise} 
In our experiment, we observed that financial data contains significant noise, which complicates the training process. Although we attempted to alleviate this through normalization (Eq. \ref{norm}), the performance improvement was minimal. To address this challenge, we consider two approaches. One is to improve the model structure. For instance, \cite{piovesan2023power} suggests using a Gaussian distribution for regression tasks instead of predicting a single value, which has proven effective in high-noise or dynamic data scenarios. The other approach involves employing feature engineering methods. For example, we identified some effective features, as shown in Fig. \ref{fig:factor}, which may help enhance model robustness and performance.

\subsection{Domain Shift}
During the project, we mistakenly used the mean and standard deviation of the entire dataset for normalization instead of using a rolling one-minute window. This oversight led to a noticeable improvement in model performance compared to the current results. This suggests that the testing data may exhibit a significant domain gap from the training data, indicating that the data domain gradually changes over time. To address this problem, we may need to explore cross-domain methods \cite{zhao2024knn, zhao2024crossfi}.

\subsection{Limitations}
This work also has several limitations that can be explored in future research. First, regarding backbone model selection, the currently chosen models can be categorized into MLP and decoder structures (LSTM and Mamba). However, encoder structures are also important in time series analysis. We have developed a BERT model for financial data (following the approach in \cite{zhao2024mining}) as provided in the repository. Due to time constraints, however, we have not fully tested its performance. Second, our current approach to addressing label imbalance primarily focuses on loss functions. Other methods, such as data augmentation and few-shot learning, could also be explored. Third, if data and resources permit, we could attempt to construct a large foundation model for financial data \cite{chen2024overview}. As mentioned in the last section, we found that model performance was significantly poor when we trained the model on a single item, but this issue was resolved when we used multiple items. We believe that scaling laws can also be beneficial in the field of finance.

\section{Conclusion}
In this project, we attempt to learn the prediction for the forward 1 min return in the Chinese future market. We succeed in addressing the substantial challenges in high-frequency trading nature and building a model with the stable predictive power by proper backbone models and label imbalance adjustment methods.

\bibliographystyle{ieeetr}
\bibliography{ref.bib}

\end{document}